\documentclass[runningheads]{llncs}
\usepackage{graphicx}
\usepackage{hyperref}
\begin{document}
\title{ICodeNet - A Hierarchical Neural Network Approach for Source Code Author Identification\thanks{Supported by L3Cube Pune.}}
\titlerunning{ICodeNet - A Hierarchical Approach for Source Code Author Identification}
%
\author{Pranali Bora\inst{1} \and
Tulika Awalgaonkar\inst{1} \and
Himanshu Palve\inst{1} \and
Raviraj Joshi\inst{2} \and
Purvi Goel\inst{2}}
\authorrunning{P. Bora et al.}
%
\institute{Pune Institute of Computer Technology \and
Indian Institute of Technology Madras\\
\email{\{pranalibora98, tulika.awalgaonkar, himanshupalve1999\}@gmail.com}\\
\email{\{ravirajoshi,goyalpoorvi\}@gmail.com}}
\maketitle              
\begin{abstract}
With the open-source revolution, source codes are now more easily accessible than ever. This has, however, made it easier for malicious users and institutions to copy the code without giving regards to the license, or credit to the original author. Therefore, source code author identification is a critical task with paramount importance. In this paper, we propose ICodeNet - a hierarchical neural network that can be used for source code file-level tasks. The ICodeNet processes source code in image format and is employed for the task of per file author identification. The ICodeNet consists of an ImageNet trained VGG encoder followed by a shallow neural network. The shallow network is based either on CNN or LSTM. Different variations of models are evaluated on a source code author classification dataset. We have also compared our image-based hierarchical neural network model with simple image-based CNN architecture and text-based CNN and LSTM models to highlight its novelty and efficiency.

\keywords{Author Identification \and Convolutional Neural Network \and Long Short Term Memory \and Transfer Learning.}
\end{abstract}
\section{Introduction}
As the amount of publicly available source code increases, even the volume and diversity of software programs available to the software community increases. A large number of sources available online are unknown and often lack authorship. Therefore it is important to develop mining techniques for understanding code and related authors. It is one of the main attributes for measuring the contribution and importance of researchers and has important academic, social, and financial implications.  
\par Recently, the code authorship identification task has gained increased awareness in the software industry due to its application. The main application includes code plagiarism detection\cite{Burrows2007EfficientPD}, automated code generation, cyber-attack investigation, code integrity investigations\cite{forensics2008investigating}, and malicious code detection \cite{krsul1997authorship}. Code Authorship Identification is the task of identifying the programmer of a particular piece of code. Every programmer has a unique and distinct style of writing the code known as code stylometry. Style is based on various factors such as the way of defining variables, indentation and commenting styles, and the number of lines per function. 
\par The Author identification task is commonly done using Natural Language Processing (NLP) techniques that would successfully derive semantic and syntactic features for classification. This process would generally proceed by tokenization and vocabulary creation. Optionally it may involve removing stop words, casings, and reducing words to their stem versions. The NLP techniques are mostly augmented with manual features as it is not possible to capture important information using textual analysis. The features like average variable length, type of variable casing, indentation style, the ratio of comment lines to code lines, white spaces per line, and formatting of operators are unique to the authors and hence very helpful in the author identification task \cite{yang2017authorship,ding2002extraction,lim2009method}. Most of these features have to be explicitly captured and cannot be automatically captured using text-based approaches. This work is in the direction of getting rid of manual feature engineering completely. Our approach aims to capture the structural features of code that are unique to a particular author while eliminating preprocessing overheads of tokenization. 
\par In this work, we aim at solving the task of author identification using its visual features. Our work makes use of the SnapCode\cite{9031980} approach which introduces the idea of extracting structural and syntactic features automatically from the source code using image-based processing techniques. The SnapCode approach completely eliminates the tasks of manual feature engineering and tokenization while preserving the unique features within the source code. It does so by creating an image of the code snippet which can be visualized as taking a snapshot of the snippet. The snapshot contains the entire code snippet along with its layout thus eliminating loss of any information or feature. This snapshot is directly processed by the neural network allowing it to automatically learn the discriminative features. We use a similar approach for representing the source code snippets as code snapshots. 
\par The entire source code file can be seen as a sequence of code snippets. So the problem of processing a source code file can be formulated as a sequence processing task. Each element of this sequence is a code snippet. Since we process images instead of text, each element is a code snapshot. This formulation allows us to use a deep neural network to first extract representations of these individual snapshots. The sequence of representations can be fed to another neural network for final classification. We thus propose ICodeNet which employs hierarchical neural networks for the task of author identification. This hierarchical setup helps in increasing the accuracy of author identification to a great extent. This work is concerned with the file-level classification of source codes as opposed to the fragment-level classification. File-level or document level classification is desirable when the entire file is written by a single author. Moreover, it may sometimes become hard to identify the author from the individual small code snippets which mostly look alike across different authors, therefore looking at the entire file may be desirable. With this motivation, we explore file-level classification utilizing the SnapCode approach. This is the first work to employ an image-based technique to document level author attribution. Although ICodeNet is applied to the task of author identification for source code files, it is generic enough to model any file-level task for text data. It can also be used to model classification tasks related to very long text sequences. 
\par The main contributions of this paper are as follows: 
\begin{itemize}
    \item We introduce hierarchical neural networks to encode source code file information and show that it captures representations helpful for the task of author identification.
    \item We show that transfer learning from ImageNet can also work in the context of source code. We propose a simple fine-tuning approach to adapt a pre-trained VGG network to source code snippet images.
    \item We propose data augmentation techniques specific to our use case which allows us to train the model even with very few source files for an author.
\end{itemize}

\section{Background}

\subsection{Related Work}

The classic literature authorship identification problem has largely motivated and inspired research in the field of source code authorship identification. Even though natural languages have more flexible grammatical rules than programming languages, there is a large degree of variation in the code writing styles of various programmers. For example, the coding styles exhibited by experienced programmers are different than those of the novice ones. Different approaches have been applied for automatic author identification to date. Early work in this domain used plain text features of source code archives to identify the authors. Frantzeskou \textit{et.al}. in their work\cite{frantzeskou2007identifying} used a technique to extract the frequency of sequences of n-characters from the source code files to form an author profile which was compared with the test author files. Analysis of code depends on the sequences of text occurring in the source code. One of the widely used methods to preprocess text data includes tokenization, separation of alphanumeric characters and symbols, removal of unknowns, replacement of tokens by indexes in the dictionary, and padding. After the completion of preprocessing and feature extraction various machine learning algorithms like Naive Bayes, Support Vector Machines, k-nearest neighbor, Random Forest classifier, etc. are applied to classify the source code depending on their authors \cite{7881396}. 
\par One of the most successfully used ways for representing source code is Abstract Syntax Trees(AST). AST’s have also been used in code clone detection because of their ability to capture structural features of the code. Lazar \textit{et.al}. in their work\cite{6840038}, extract an AST for each program by parsing the source code. Hash codes are then created for each subtree and provided as input to the clone detection algorithms. However, AST’s drop the formatting features and rely on hand-constructed features which may be obfuscated. 
\par Previous studies used various types of features to improve the accuracy of the classification and implemented a feature selection technique to remove the least significant features\cite{6664734,elenbogen2008detecting}. An unsupervised approach for source code attribution was proposed by Bandara \textit{et.al}.\cite{bandara2013source}. Features like the number of characters and words per line, frequency of access specifiers and comments, length of each identifier, number of underscores, etc were manually extracted and a sparse auto-encoder was used to learn a non-linear representation of these features. These features were used as inputs for the logistic regression model which was built for every author. 
\par Further progress into the domain saw many deep learning architectures being used depending on the various applications. These architectures eliminated the task of manual feature engineering and extracted the features automatically from the source code\cite{zekany2016crystalball}. Convolutional neural network(CNN) is one such deep learning architecture that is very popular these days. CNNs have shown that they can achieve excellent results in many natural language processing tasks such as search query retrieval\cite{shen2014learning}, text summarization\cite{alquliti2019summarization}, semantic parsing\cite{yih2014semantic}, sentence modeling\cite{kalchbrenner2014convolutional} etc. In this work, we propose an architecture that uses the capabilities of CNNs to extract discriminative features automatically from the source code archives and attributes them to the different authors. In a recent work\cite{abuhamad2019code}, the source code samples were represented using two techniques, namely term frequency-inverse document frequency(TFIDF) and word embeddings before feeding them as inputs to the CNN model. The TFIDF approach reflected the importance of a word to a document in the collection, and the word embeddings approach allowed similar representations for words with similar meanings. However, our approach is based on images and helps us better represent structural aspects of the code which otherwise can only be captured manually.  In this work, we use both the CNN and LSTM architectures for the classification of authors.

\subsection{Convolutional Neural Networks}

A convolutional neural network is a type of deep neural network that is mostly used to analyze the visual data \cite{krizhevsky2012imagenet}. A deep ConvNet consists of multiple hidden layers which include the convolutional, pooling, non-linearity, batch normalization, and fully connected layers. It is mainly used in the field of image recognition, anomaly detection, drug discovery, video analysis, etc. A ConvNet has the ability to learn the features of the input which were otherwise hand-engineered in the primitive methods. Each layer in the ConvNet learns different features during training, with increasing complexity as the network gets deeper. As CNNs can successfully capture spatial and temporal dependencies in an image, they can be employed to solve challenging problems like plagiarism detection, author identification, malware detection, and code analysis in a more efficient manner.

\subsection{Long Short Term Memory}

A long short term memory is an improvement over the RNNs as they are capable of storing information for a longer period of time \cite{hochreiter1997long}. They are widely used in the field of text recognition, language translation, image captioning because their predictions are conditioned by the past experience of the network’s inputs. LSTMs have the ability to successfully learn on data with long-range temporal dependencies. So they are mainly used for sequence processing applications. The LSTM architecture consists of memory blocks/cells and mechanisms called gates to perform manipulation operations on the cells. These gates learn the relevance of information and decide the information to be kept or forgotten during training. LSTMs have been long used in many NLP tasks to give remarkable results. In this work, the sequence of vectoral representation of code snapshots is modeled using LSTMs.



\subsection{Transfer Learning}

Transfer learning is the improvement in learning a new task through the transfer of knowledge from a related task that has already been learned. It is a popular approach in deep learning where parameters of the model trained on a large dataset can be reused to solve a related problem. The extracted weights and features of the state-of-the-art pre-trained models are used for the new task to improve its performance. Transfer learning strategies can be applied to a wide multitude of domains like text classification, spam filtering, online recommendations, and medical applications\cite{raghu2019transfusion}. 
\par In this work, we have used transfer learning from the ImageNet dataset by utilizing the pre-trained VGG16 model. The VGG16 model pre-trained on the ImageNet dataset\cite{simonyan2014very} is used as a feature extractor and is also finetuned on the target dataset. VGG16 is a deep CNN architecture with 16 weight layers. It is one of the most popular pre-trained models used today.

\section{Approach}

Tokenization is elementary in all text processing applications. It aims at splitting the sentence into a stream of tokens followed by using processing methods like stemming and lemmatization. This type of preprocessing eliminates features like indentation information, number of characters per line, length of identifiers, and the way of defining a function that is essential to the task at hand. With the current approaches, the only way to extract these features is by hand-engineering. 
\par Therefore, the ICodeNet carries out automatic extraction of features for the task of author identification using image processing techniques in a hierarchical setup. 

\subsection{Dataset}

The dataset used was raw source code Java files taken from the GitHub repositories of various authors. The original dataset consisted of source code archives for 40 different authors \cite{yang2017authorship}. The dataset was highly imbalanced with some authors having very few source code files. So, the authors having less than 50 files were removed from the dataset. Thus, the final dataset consisted of the raw source code Java files for 20 different authors. The data is shared publicly to enable further research in this area \footnote{\href{https://github.com/l3cube-pune/ICodeNet}{Author Identification Dataset}}.
\par This dataset is a good representative of real-world code examples, as opposed to the competitive coding examples, used by most of the other works \cite{simko2018recognizing,alsulami2017source,caliskan2015anonymizing,abuhamad2019code}. The problems with large competitive coding datasets that give near-perfect accuracy are thoroughly discussed in \cite{bogomolov2020authorship}. The author name at the top of the source code files was explicitly removed.

\subsection{Preprocessing}

\begin{enumerate}
    \item \textbf{Splitting the Data}.  
    The dataset was split into two subsets - train and test. The test set consisted of 30 files per author and the remaining files were added to the train set. The average number of train files was 57 with an average of 100 lines of code per file. The data regarding the total number of java files and the average number of lines of code per file is as shown in Fig. ~\ref{fig:fig1}. 
    
    \begin{figure}[!htbp]
    \centering
    \includegraphics[scale=0.4]{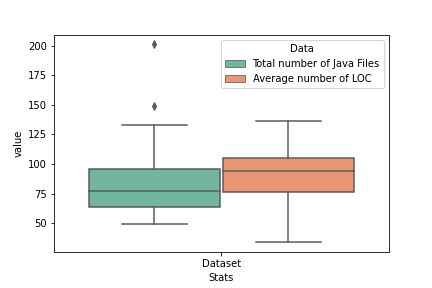} 
    \caption{Dataset Details.}
    \label{fig:fig1}
    \end{figure}
    
    \item \textbf{Creating the Chunks}. As the number of lines of code per file for the authors is varying, the source code files were divided into partially overlapping snippets of 20 lines each with a shift of 5 lines for the successive chunk. The chunks were created using the grouper function provided by python. Oversampling was done on this overlapped data to balance the data of different authors. It was ensured that the train and test set do not contain snippets from the same file. Therefore, we propose a simple data augmentation technique by varying the number of lines in a chunk and the shift size. This allows us to repeat the same file having different chunk and shift sizes for each repetition. Oversampling and overlapping techniques were only used for training the encoder network. Encoded representations of non-overlapping chunks were fed to the shallow network. 
    
    \item \textbf{Cleaning the Chunks}. The chunks so formed, were cleaned to get useful data for classification. All the chunks with less than 20 lines of code and fewer(less than 5) characters per line were discarded to get a cleaner version of the data. The chunks per author in the train set were equalized to 1000 to ensure no imbalance in the data. The chunks were equalized for fine-tuning the VGG network explained in the next section.
    
    \item \textbf{Image Formation}. The cleaned, overlapped, and oversampled chunks of data were converted into images with a resolution of 256x256 pixels using the Pillow API. These images were fed to the VGG encoder for classification.

\end{enumerate}

\begin{figure}[!htbp]
    \centering
    \includegraphics[scale=0.5]{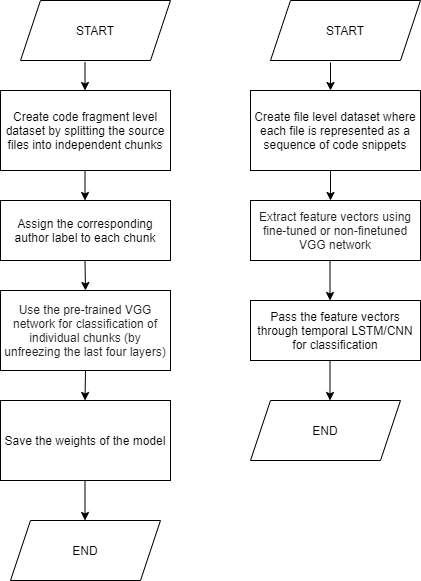} 
    \caption{Flowchart (left) describes the process of fine-tuning the VGG network on individual snapshot or fragment classification task (auxiliary task) and the flowchart (right) describes the entire process of classification of source code files into their respective authors (original task).}
    \label{fig:fig4}
    \end{figure}

\subsection{Classification}
After pre-processing each source code file is represented as a sequence of images. The process of classification is carried out in two steps. In the initial step, individual images are encoded into a fixed dimensional vector. This can also be seen as converting raw images into feature vectors. The sequence of feature vectors is then used for classification. 
\par A VGG16 model was used for extracting features from the image snapshots. For a sequence of snapshots, each individual snapshot was passed through the VGG16 model, and the resulting feature vectors were passed through another neural network for classification. The VGG network was used because it is one of the popular models for feature extraction in vision tasks \cite{canziani2016analysis}. The pre-trained VGG network was also separately finetuned on an auxiliary source code image classification task as it was originally pre-trained on conventional images. The initial layers of VGG were frozen and the last four layers were fine-tuned. An auxiliary dataset of code snapshot images and their authors was created by splitting the parent file and converting it to images. The VGG network was then trained on the task of classifying individual snapshots by adding a task-specific dense layer. This task can be seen as a simple image classification task where individual snapshots of the source file are training samples and labels are the corresponding author names. As opposed to the parent task of author identification for a complete file this finetuning subtask identifies the author for code fragments. This helped the VGG network to better adapt to the target dataset. The convolutional layers of this VGG network were used as a feature extractor. Note that instead of training and finetuning in an end to end network, the VGG was separately finetuned on a similar classification task. The original pre-trained VGG is referred to as non-finetuned VGG and the VGG re-trained on source code snapshots is referred to as fine-tuned VGG in this work. The complete flow is described in Fig. ~\ref{fig:fig4}.
\par A total of 20000 training samples and 13263 testing samples, created using the overlapping and oversampling techniques. The model was trained using the RMSProp optimizer and categorical cross-entropy loss function. After training, feature vectors were extracted from the 1024 neuron dense layer. The extracted feature vectors represented the individual chunks. As multiple chunks were created per file, the feature vectors of the chunks belonging to the same file were concatenated in order, to give feature vectors for the entire file. The feature vectors obtained per file had variable length, as the number of chunks per file was variable. A different number of chunks per file were created due to the variable number of lines of code per file. 
\par The extracted feature vectors were used in two settings. In the first approach, the VGG was followed by a 1D-CNN based network, we represent it as VGG + CNN. The second approach used LSTM instead of a CNN to process the time series of feature vectors and is represented as VGG + LSTM. One of the reasons for having this hierarchical structure is the large size of the input files. It was difficult to train a VGG only network with a single big image file for the entire source code file. So the snippet based approach is used where the features obtained from individual snippets are then merged using CNN and LSTM based networks. 
\par
\begin{figure}
    \centering
    \includegraphics[scale=0.25]{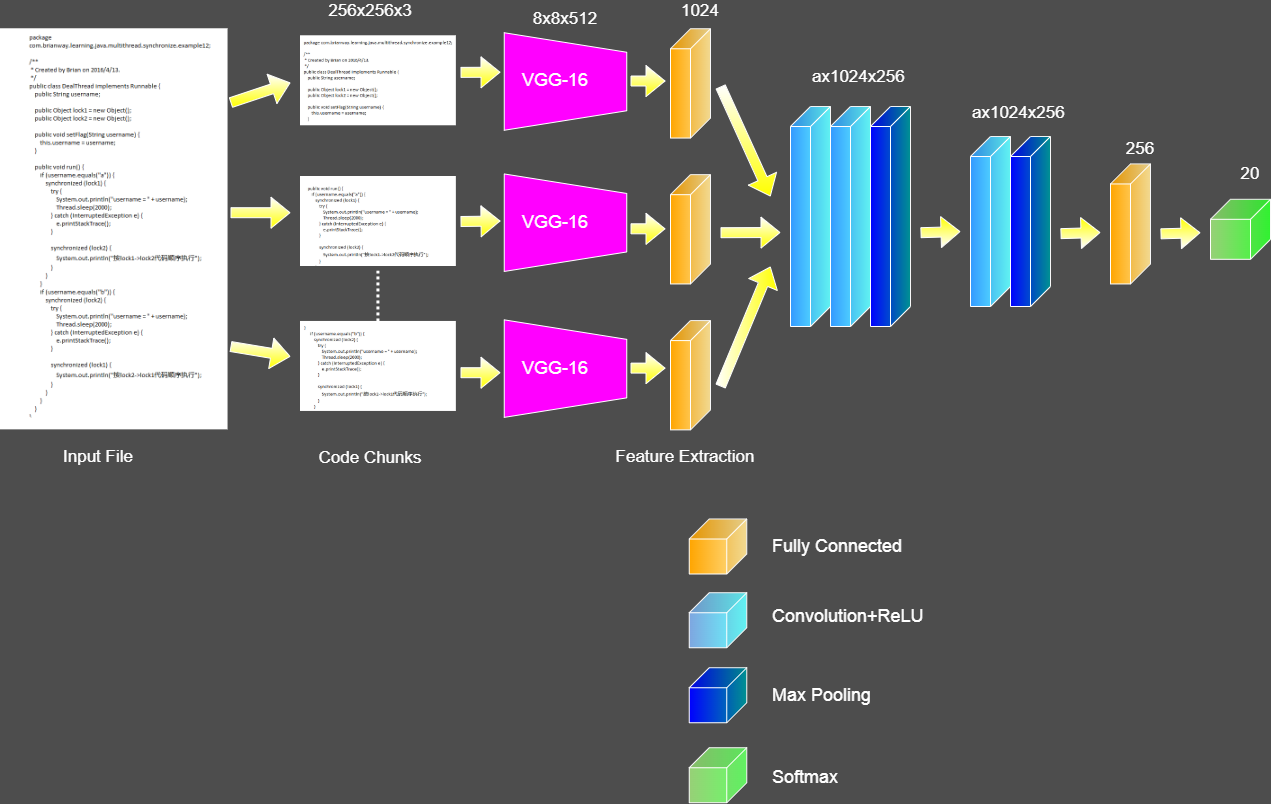} 
    \caption{VGG+CNN Architecture.}
    \label{fig:fig2}
    \end{figure}

\begin{figure}
    \centering
    \includegraphics[scale=0.25]{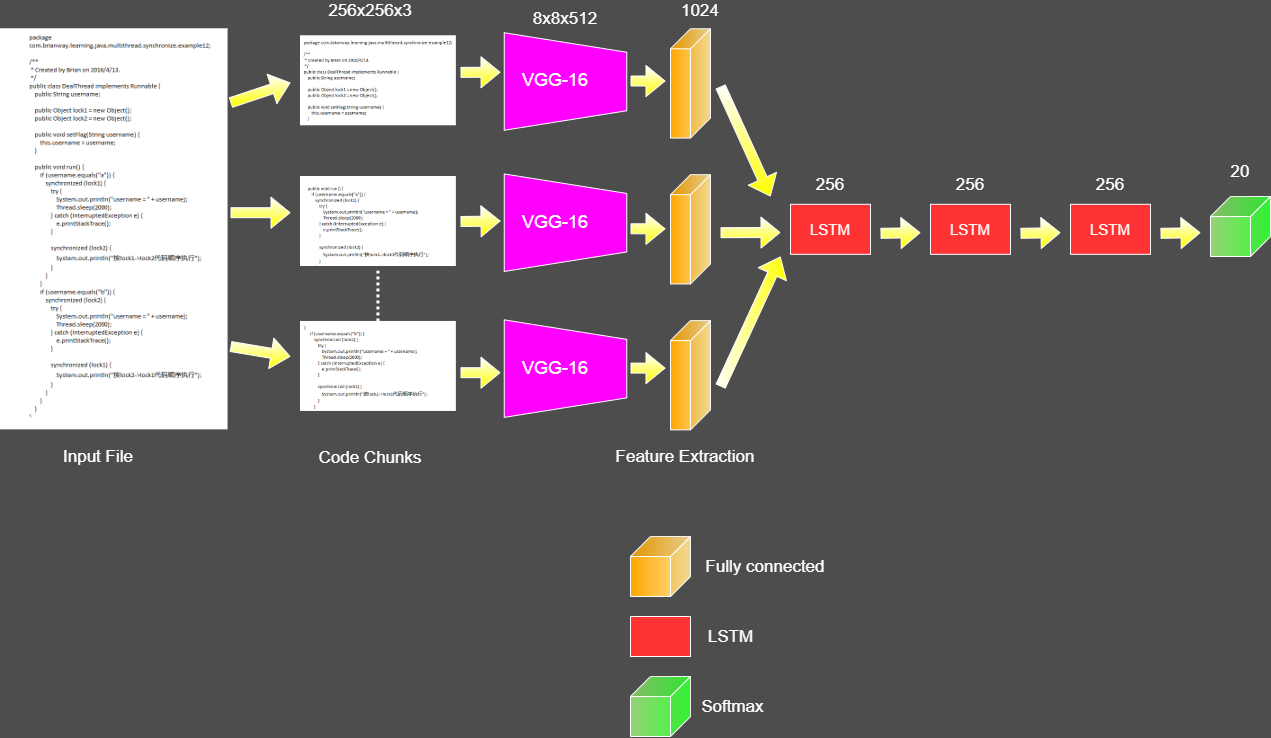} 
    \caption{VGG+LSTM Architecture.}
    \label{fig:fig3}
    \end{figure}

 1D-CNN\cite{kiranyaz20191d} is very effective for the detection of features that span over multiple segments of the dataset. The 1D-CNN model consists of three 1D-Convolutional layers with 256 filters of size 3 followed by two dense layers. The network architecture is shown in Fig.~\ref{fig:fig2}. 
In the second approach, LSTM was used for the classification task. As shown in Fig.~\ref{fig:fig3}, three LSTM layers with 256 units are stacked together to form the network. Feature vectors extracted from the fine-tuned VGG model were fed as inputs to both models. In both approaches, a softmax layer was used for classification. Therefore, the entire file classification is done using these two novel architectures utilizing the hierarchical setup naturally presented by the data. This separately fine-tuned VGG encoder followed by a shallow time-series network is termed the ICodeNet Architecture.

\section{Results}
Along with VGG + CNN and VGG + LSTM, the results for VGG + Voting are also evaluated for comparison. VGG + Voting refers to the classification of individual chunks using VGG followed by a voting layer to determine the label for the entire file. Another model used for comparison is a simple CNN(s-CNN) based image encoder instead of the VGG encoder. The s-CNN model consists of five alternating convolutional and max-pooling layers with two dense layers stacked on top for feature extraction. Similar to the VGG encoder, this model was also separately trained on snapshot data and then used as a feature extractor. 

\begin{table}
\centering
\caption{Classification accuracies over different architectures}
\label{tab:tab1}       
%
%
\begin{tabular}{p{5cm}p{2cm}p{2cm}p{2cm}}
\hline\noalign{\smallskip}
\textbf{Neural Network architecture} & \textbf{Fine-tuned VGG} & \textbf{Fine-tuned VGG} & \textbf{S-CNN}  \\ \hline
\noalign{\smallskip}\noalign{\smallskip}
Image Encoder + Voting & 73.90\% & 72.00\% & 66.98\% \\ \hline
\noalign{\smallskip}\noalign{\smallskip}

Image Encoder + CNN    & 82.88\% & 79.02\% & 69.32\% \\ \hline
\noalign{\smallskip}\noalign{\smallskip}

Image Encoder + LSTM   & \textbf{85.42}\% & 82.04\% & 72.00\% \\ \hline
\noalign{\smallskip}\noalign{\smallskip}

\end{tabular}
\end{table}

\par
\begin{figure}[!htbp]
    \centering
    \includegraphics[width=60mm]{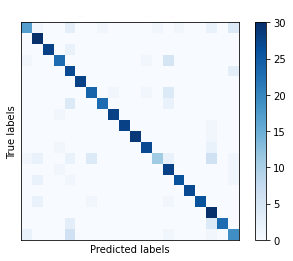} 
    \caption{Confusion matrix of the best accuracy model.}
    \label{fig:fig5}
\end{figure}

 Both the steps i.e. feature extraction and classification using hierarchical networks were carried out on the two models. The fine-tuned VGG based CNN model gave an accuracy of 73.90\% after voting as compared to 72\% by the non-fine tuned model and 66.98\% by s-CNN. The drop in accuracy of s-CNN shows that it was not able to properly identify the discriminative features from code snippets like the VGG based model. The feature vectors were extracted from both these models to perform the further tasks. An accuracy of 82.88\% was reported with the fine-tuned VGG+CNN model, whereas the non-fine tuned VGG+CNN model reported a slightly less accuracy of 79.02\%. With, s-CNN+CNN the accuracy further dropped to 69.32\%. Similarly, the same set of tasks was performed for the hierarchical LSTM model, and results were observed. The fine-tuned VGG+LSTM architecture gave the best accuracy of 85.42\%. The confusion matrix for this best configuration is shown in Fig.~\ref{fig:fig5}. It indicates that the accuracy is not biased towards some specific authors. While the non-fine tuned VGG+LSTM model resulted in 82.04\% accuracy and s-CNN+LSTM gave an accuracy of 72\% with the same dataset.

The results have been combined in Table~\ref{tab:tab1}. The results show that fine-tuning helps in adjusting the parameters of the model very precisely thus, improving its performance to give the most optimized outcomes. They also indicate that a simple CNN model cannot be used to represent the discriminative features and style of writing the code with the same effectiveness as our models. 
\par We have also compared our approach with standard text-based CNN and LSTM models to further highlight its efficiency. The LSTM model architecture consisting of two layers of bidirectional LSTM with 256 and 128 neurons respectively followed by a global max-pooling layer and dense layer with 128 neurons gave an accuracy of 74.83\%. The text-based CNN model with two 1D-CNN layers of 256 and 128 filters respectively with a filter size of 4 followed by a global max-pooling layer and dense layer with 128 neurons gave an accuracy of 81.33\%. 
\par The ICodeNet approach performs better than text-based approaches. It was able to preserve the features and code styling as required for author classification. These features would rather have been very difficult to extract with the traditional methods. 

\section*{Conclusion}
In this work, we proposed a methodology to extract style-related features from source code automatically, which otherwise have been highly difficult to define and extract. We have used two hierarchical neural networks for the classification of the extracted feature representations according to their authors. The pre-trained CNN model based on the VGGNet was used to extract discriminative features from the source code snapshots which were later fed to two different neural networks i.e CNN and LSTM for classification. We assume that each file is specific to an author. Handling of multi-author source codes is left to future scope.

\section*{Acknowledgements} This work was done under the L3Cube Pune mentorship program. We would like to express our gratitude towards our mentors at L3Cube for their continuous support and encouragement.
%
%
%
%
\bibliographystyle{splncs04}
\bibliography{main}

\begin{thebibliography}{10}
\providecommand{\url}[1]{\texttt{#1}}
\providecommand{\urlprefix}{URL }
\providecommand{\doi}[1]{https://doi.org/#1}

\bibitem{abuhamad2019code}
Abuhamad, M., Rhim, J.s., AbuHmed, T., Ullah, S., Kang, S., Nyang, D.: Code
  authorship identification using convolutional neural networks. Future
  Generation Computer Systems  \textbf{95},  104--115 (2019)

\bibitem{alquliti2019summarization}
Alquliti, W., Binti, N.: Convolutional neural network based for automatic text
  summarization. International Journal of Advanced Computer Science and
  Applications  \textbf{10} (01 2019). \doi{10.14569/IJACSA.2019.0100424}

\bibitem{alsulami2017source}
Alsulami, B., Dauber, E., Harang, R., Mancoridis, S., Greenstadt, R.: Source
  code authorship attribution using long short-term memory based networks. In:
  European Symposium on Research in Computer Security. pp. 65--82. Springer
  (2017)

\bibitem{bandara2013source}
Bandara, U., Wijayarathna, G.: Source code author identification with
  unsupervised feature learning. Pattern Recognition Letters  \textbf{34}(3),
  330--334 (2013)

\bibitem{bogomolov2020authorship}
Bogomolov, E., Kovalenko, V., Bacchelli, A., Bryksin, T.: Authorship
  attribution of source code: A language-agnostic approach and applicability in
  software engineering. arXiv preprint arXiv:2001.11593  (2020)

\bibitem{Burrows2007EfficientPD}
Burrows, S., Tahaghoghi, S.M.M., Zobel, J.: Efficient plagiarism detection for
  large code repositories. Softw. Pract. Exp.  \textbf{37},  151--175 (2007)

\bibitem{caliskan2015anonymizing}
Caliskan-Islam, A., Harang, R., Liu, A., Narayanan, A., Voss, C., Yamaguchi,
  F., Greenstadt, R.: De-anonymizing programmers via code stylometry. In: 24th
  $\{$USENIX$\}$ Security Symposium ($\{$USENIX$\}$ Security 15). pp. 255--270
  (2015)

\bibitem{canziani2016analysis}
Canziani, A., Paszke, A., Culurciello, E.: An analysis of deep neural network
  models for practical applications. arXiv preprint arXiv:1605.07678  (2016)

\bibitem{ding2002extraction}
Ding, H.: Extraction of Java Program Fingerprints for Software Authorship
  Identification. Ph.D. thesis, Oklahoma State University (2002)

\bibitem{elenbogen2008detecting}
Elenbogen, B.S., Seliya, N.: Detecting outsourced student programming
  assignments. Journal of Computing Sciences in Colleges  \textbf{23}(3),
  50--57 (2008)

\bibitem{forensics2008investigating}
Forensics, C.H.M.M.: Investigating and analyzing malicious code. Cameron H.
  Malin, Eoghan Casey, James M. Aquilina.--1 edition.--Waltham: Syngress
  (2008)

\bibitem{frantzeskou2007identifying}
Frantzeskou, G., Stamatatos, E., Gritzalis, S., Chaski, C.E., Howald, B.S.:
  Identifying authorship by byte-level n-grams: The source code author profile
  (scap) method. International Journal of Digital Evidence  \textbf{6}(1),
  1--18 (2007)

\bibitem{hochreiter1997long}
Hochreiter, S., Schmidhuber, J.: Long short-term memory. Neural computation
  \textbf{9}(8),  1735--1780 (1997)

\bibitem{kalchbrenner2014convolutional}
Kalchbrenner, N., Grefenstette, E., Blunsom, P.: A convolutional neural network
  for modelling sentences. arXiv preprint arXiv:1404.2188  (2014)

\bibitem{kiranyaz20191d}
Kiranyaz, S., Avci, O., Abdeljaber, O., Ince, T., Gabbouj, M., Inman, D.J.: 1d
  convolutional neural networks and applications: A survey. arXiv preprint
  arXiv:1905.03554  (2019)

\bibitem{krizhevsky2012imagenet}
Krizhevsky, A., Sutskever, I., Hinton, G.E.: Imagenet classification with deep
  convolutional neural networks. Advances in neural information processing
  systems  \textbf{25},  1097--1105 (2012)

\bibitem{krsul1997authorship}
Krsul, I., Spafford, E.H.: Authorship analysis: Identifying the author of a
  program. Computers \& Security  \textbf{16}(3),  233--257 (1997)

\bibitem{6840038}
{Lazar}, F., {Banias}, O.: Clone detection algorithm based on the abstract
  syntax tree approach. In: 2014 IEEE 9th IEEE International Symposium on
  Applied Computational Intelligence and Informatics (SACI). pp. 73--78 (2014)

\bibitem{lim2009method}
Lim, H.i., Park, H., Choi, S., Han, T.: A method for detecting the theft of
  java programs through analysis of the control flow information. Information
  and Software Technology  \textbf{51}(9),  1338--1350 (2009)

\bibitem{raghu2019transfusion}
Raghu, M., Zhang, C., Kleinberg, J., Bengio, S.: Transfusion: Understanding
  transfer learning for medical imaging. In: Advances in neural information
  processing systems. pp. 3347--3357 (2019)

\bibitem{7881396}
{Reyes}, J., {Ramírez}, D., {Paciello}, J.: Automatic classification of source
  code archives by programming language: A deep learning approach. In: 2016
  International Conference on Computational Science and Computational
  Intelligence (CSCI). pp. 514--519 (2016)

\bibitem{9031980}
{Sarnot}, S.A.P., {Rinke}, S., {Raimalwalla}, R., {Joshi}, R., {Khengare}, R.,
  {Goel}, P.: Snapcode - a snapshot based approach to code stylometry. In: 2019
  International Conference on Information Technology (ICIT). pp. 337--341
  (2019)

\bibitem{shen2014learning}
Shen, Y., He, X., Gao, J., Deng, L., Mesnil, G.: Learning semantic
  representations using convolutional neural networks for web search. In:
  Proceedings of the 23rd international conference on world wide web. pp.
  373--374 (2014)

\bibitem{simko2018recognizing}
Simko, L., Zettlemoyer, L., Kohno, T.: Recognizing and imitating programmer
  style: Adversaries in program authorship attribution. Proceedings on Privacy
  Enhancing Technologies  \textbf{2018}(1),  127--144 (2018)

\bibitem{simonyan2014very}
Simonyan, K., Zisserman, A.: Very deep convolutional networks for large-scale
  image recognition. arXiv preprint arXiv:1409.1556  (2014)

\bibitem{6664734}
{Tennyson}, M.F.: A replicated comparative study of source code authorship
  attribution. In: 2013 3rd International Workshop on Replication in Empirical
  Software Engineering Research. pp. 76--83 (2013)

\bibitem{yang2017authorship}
Yang, X., Xu, G., Li, Q., Guo, Y., Zhang, M.: Authorship attribution of source
  code by using back propagation neural network based on particle swarm
  optimization. PloS one  \textbf{12}(11),  e0187204 (2017)

\bibitem{yih2014semantic}
Yih, W.t., He, X., Meek, C.: Semantic parsing for single-relation question
  answering. In: Proceedings of the 52nd Annual Meeting of the Association for
  Computational Linguistics (Volume 2: Short Papers). pp. 643--648 (2014)

\bibitem{zekany2016crystalball}
Zekany, S., Rings, D., Harada, N., Laurenzano, M.A., Tang, L., Mars, J.:
  Crystalball: Statically analyzing runtime behavior via deep sequence
  learning. In: 2016 49th Annual IEEE/ACM International Symposium on
  Microarchitecture (MICRO). pp. 1--12. IEEE (2016)

\end{thebibliography}
\end{document}